\pdfoutput=1
\documentclass{article}
\usepackage{iclr2027_conference,times}

\usepackage[utf8]{inputenc}
\usepackage[T1]{fontenc}
\usepackage{hyperref}
\usepackage{url}
\usepackage{booktabs}
\usepackage{tabularx}
\usepackage{multirow}
\usepackage{nicefrac}
\usepackage{amsmath,amssymb,amsthm}
\usepackage{wrapfig}
\usepackage{graphicx}
\graphicspath{{figs/}}
\usepackage{xcolor}
\usepackage{tikz}
\usetikzlibrary{arrows.meta,positioning,calc,fit,backgrounds}
\usepackage{algorithm}
\usepackage{algpseudocode}

\usepackage{makecell}
\usepackage{placeins}
\usepackage{float}

\usepackage{multirow}
\usepackage{graphicx}

\theoremstyle{plain}
\theoremstyle{definition}
\theoremstyle{remark}
\newcommand{\best}[1]{\textbf{#1}}
\newcommand{\second}[1]{\underline{#1}}
\newcommand{\ours}{FluidRain}

\title{FluidRain: Incompressible Rain Flow\\as an Attention Bias for Loop-in-Loop\\Video Deraining}

\author{Pu Wang$^{1}$, Yongcong Wang$^{2}$, Wenhao Li$^{1}$, Xiang Chen$^{3}$, Guangwei Gao$^{3}$, \\
\textbf{Jinshan Pan$^{3}$, Siyuan Yao$^{4}$, Shujun Fu$^{1}$, Zhuoran Zheng$^{5}$} \\
$^{1}$Shandong University \quad $^{2}$Central South University \\
$^{3}$Nanjing University of Science and Technology \\
$^{4}$Beijing University of Posts and Telecommunications \\
$^{5}$National University of Defense Technology}

\iclrfinalcopy
\begin{document}
\maketitle
\lhead{Preprint}

\begin{abstract}
Existing video deraining methods typically exploit neighboring frames through either explicit alignment or implicit spatiotemporal aggregation.     Explicit alignment relies on accurate motion estimation, which can become unreliable under dense rain, while implicit aggregation avoids alignment but lacks explicit guidance on the directional and temporally coherent structure of rain.  
This leaves a gap between reliable temporal aggregation and explicit modeling of rain motion.

To address these limitations, we propose FluidRain, a lightweight video derainer that uses divergence-free rain flow to guide \textbf{Loop-in-Loop} attention across scales and neighboring frames.
Motivated by fluid mechanics, we model rain motion as a divergence-free image-space flow and use it to organize multi-scale and temporal aggregation.
Specifically, FluidRain first estimates a rain-flow field for each frame and projects it onto the divergence-free subspace.
The resulting flow steers window attention along rain streaks, enabling neighboring frames to be aggregated without explicit alignment.
Since rain-flow structure is preserved across scales and nearby frames, Loop-in-Loop reuses the same attention operator across both dimensions, resulting in a three-frame model with only \textbf{0.80M} parameters.
Experiments on four benchmarks show that FluidRain remains competitive with substantially larger restoration models. 
We further examine how temporal evidence scales with different input views. To evaluate whether the model remains reliable when rain motion changes across frames, we introduce RainSyn-Gust, which injects controlled changes in rain-streak direction into existing benchmarks. We also develop a physics-based no-reference metric that evaluates real-rain removal without requiring clean targets.
\end{abstract}

\section{Introduction}

Rain streaks in video arise from the motion blur of falling raindrops carried by the wind and imaged by a moving camera. Their apparent direction and length are jointly determined by wind, scene depth, and camera motion, and thus vary across sequences \citep{garg2004detection}.

Early video deraining methods were largely physics-driven. They modeled the formation and appearance of rain streaks \citep{garg2004detection,garg2006photorealistic} and combined these models with hand-crafted temporal or low-rank priors \citep{kim2015video}.
With the rise of deep learning, the field gradually shifted toward data-driven temporal modeling, first through recurrent architectures \citep{liu2018erase,yang2021recurrent,yue2021s2vd} and more recently through attention and state-space models \citep{wu2024rainmamba,sun2025vdmamba}.
Modern methods typically exploit neighboring frames in two ways. One line explicitly aligns them using optical flow or related motion estimation, whose reliability can degrade in heavily rain-corrupted regions \citep{chan2022basicvsrpp,guo2023sky}. The other aggregates neighboring features implicitly, avoiding explicit alignment but leaving the network to learn the structure of rain motion entirely from data.
Meanwhile, physical knowledge has mainly been introduced through rendering models, regularization terms, or rain synthesis \citep{yang2017jorder,hu2019dafnet,yue2021s2vd}, while geometric priors have been incorporated into attention biases \citep{liu2021swin,sun2026delivr}.
\textbf{Yet the motion structure of rain itself remains largely unused in guiding how information is aggregated across frames.}

Classical studies have shown that rain exhibits structured spatial and temporal properties rather than arbitrary corruption \citep{garg2007vision,barnum2010analysis}. Later work further characterized rain layers through attributes such as direction, scale, and temporal motion \citep{bossu2011rain,yue2021s2vd}. These regularities suggest that rain physics may provide not only guidance for feature aggregation, but also a principled basis for deciding where computation can be reused.
At the same time, looped Transformers have recently regained attention in large language models (LLMs) as a parameter-efficient alternative to conventional depth scaling. Relaxed Recursive Transformers repeatedly reuse shared blocks across depth, while recent scaling-law analyses show that recurrence can provide measurable capacity gains despite parameter sharing \citep{bae2025relaxed,schwethelm2026much}. This leads to the central question of our work: can the physical regularities of rain determine how and where a shared operator should recur in video restoration?

\begin{figure}[t]
\centering
\includegraphics[width=\linewidth]{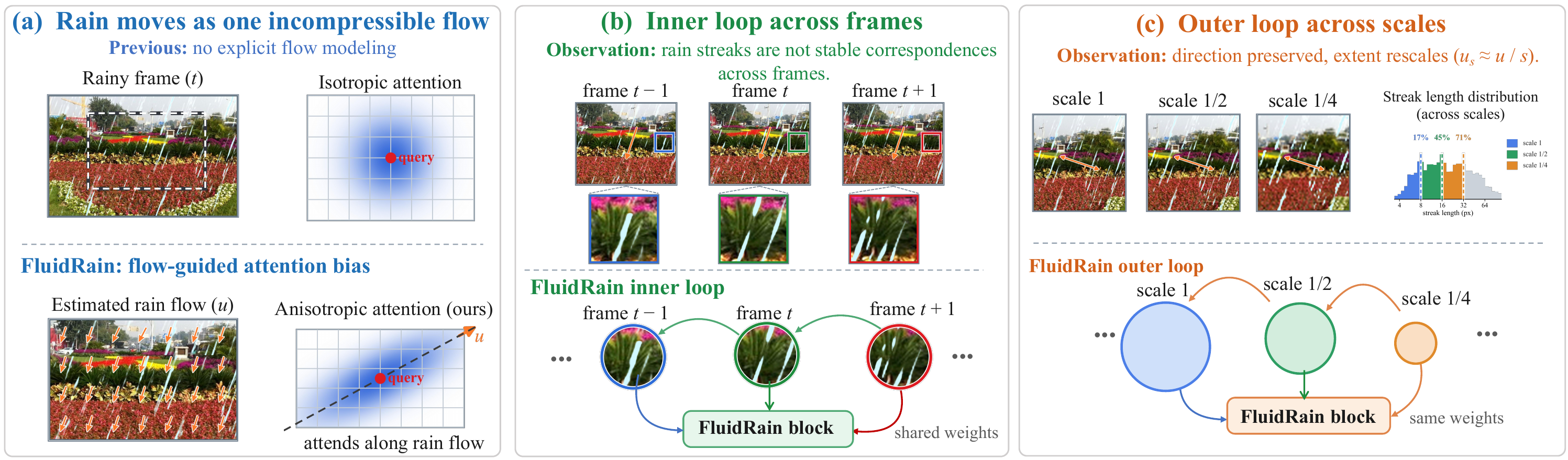}
\vspace{-5mm}
\caption{\textbf{Physical motivation and Loop-in-Loop design of FluidRain.}
(a) Rain flow guides anisotropic attention.
(b) Temporal persistence motivates the inner loop across frames.
(c) Scale consistency motivates the outer loop across resolutions.}
\vspace{-5mm}
\label{fig:teaser}
\end{figure}

To investigate this question, Figure~\ref{fig:teaser} links three physical
properties of rain to the design of FluidRain.
(a) Inspired by fluid mechanics, we model coherent rain motion within a frame
as a divergence-free image-space rain-flow field, which guides anisotropic
attention along rain streaks.
(b) Across neighboring frames, the flow remains persistent while individual
streaks are renewed, motivating an \textit{inner loop} that reuses the same
rain-aware operator over complementary observations.
(c) Across resolutions, rain-flow direction is preserved while streak extent
rescales, motivating an \textit{outer loop} that reuses the same operator
across scales.
Together, the two loops form our Loop-in-Loop design, enabling alignment-free
temporal and multi-scale aggregation with only 0.80M parameters.
Our main contributions are summarized as follows:

\begin{itemize}
\item We develop a fluid-mechanical rain-flow representation that guides attention along rain streaks, enabling temporal aggregation without relying on potentially unreliable explicit alignment under heavy rain.

\item Based on the temporal and multi-scale regularities of rain flow, Loop-in-Loop reuses the same attention operator across neighboring frames and resolutions, resulting in a 0.80M-parameter model.

\item For broader evaluation, RainSyn-Gust tests robustness to changing rain motion, while \(F_{\mathrm{phys}}\) enables no-reference assessment on real rainy videos. Extensive experiments confirm competitive restoration performance with substantially fewer parameters.
\end{itemize}

\section{Related Work}

\textbf{Video deraining.} Early video deraining methods relied on temporal filtering, low-rank priors, and recurrent architectures to exploit redundancy across neighboring frames
\citep{kim2015video,liu2018erase,yang2021recurrent,yue2021s2vd}.
With the development of attention-based models, temporal information has increasingly been aggregated through learned feature interactions rather than hand-crafted priors \citep{yang2023viwsnet}.
Recent state-space models further improve long-range spatio-temporal modeling with more favorable computational complexity.
RainMamba introduces locality-aware state-space scanning for video deraining
\citep{wu2024rainmamba}, while VDMamba combines spatial and temporal state-space branches with adaptive multi-frame fusion
\citep{sun2025vdmamba}.
Another line of work explicitly establishes temporal correspondences through optical flow, deformable alignment, or alignment-and-fusion strategies
\citep{chan2022basicvsrpp,xue2025asf}.
More recent methods have also explored semi-supervised learning, test-time adaptation, and video diffusion models to improve generalization to real-world adverse weather
\citep{lin2025controllable}.
Overall, current video deraining methods increasingly emphasize effective temporal modeling, robust multi-frame fusion, and generalization beyond synthetic rain.

\textbf{Physical priors and looped architectures.}
Physical modeling has long provided structured priors for rain removal.
Classical studies describe rain streak formation through drop motion,
exposure, and imaging geometry
\citep{garg2004detection,garg2006photorealistic,garg2007vision}.
Learning-based methods later incorporated such knowledge through rain
formation models, decomposition constraints, and learned rain synthesis
\citep{yang2017jorder,hu2019dafnet,yue2021s2vd}.
More recently, structured information has been injected directly into
feature interactions. Relative-position biases encode spatial geometry in
window attention \citep{liu2021swin}, while motion-related quantities can
guide spatio-temporal attention \citep{sun2026delivr}.
Structure can also be imposed on how computation itself is reused.
Recursive and looped networks repeatedly apply shared operators, increasing
the effective processing depth without introducing a new set of parameters
at every stage \citep{kim2016drcn,dehghani2019universal}.
This idea has recently received renewed attention in LLMs.
Relaxed Recursive Transformers reuse shared Transformer blocks across
depth \citep{bae2025relaxed}, and recent scaling analyses further study the
capacity gained from recurrence under parameter sharing
\citep{schwethelm2026much}.
Related forms of recursive reuse have also appeared in recent vision and
image-restoration models
\citep{he2026vision}.
Existing looped architectures mainly organize such recurrence along network
depth or iterative refinement.

\section{Method}
\label{sec:method}
FluidRain models rain streaks as an image-space rain-flow field, projects the
estimated flow onto the divergence-free subspace, and uses it to construct an
anisotropic attention bias. Its scale and temporal regularities further enable
the same rain-aware operator to be reused across resolutions and neighboring
frames, forming the Loop-in-Loop architecture for centre-frame restoration.

\textbf{Notation.}
Given a three-frame rainy clip
$\mathbf{I}_\tau\in\mathbb{R}^{3\times H\times W}$,
$\tau\in\{t-1,t,t+1\}$, the network predicts the restored centre frame
$\hat{\mathbf{I}}_t$, with clean target $\mathbf{I}_t^{\mathrm{gt}}$.
On synthetic benchmarks, $\mathbf{I}_\tau=\mathbf{I}_\tau^{\mathrm{gt}}+\mathbf{R}_\tau$,
where $\mathbf{R}_\tau$ is the rain layer.
Upright bold upper-case symbols denote tensors, such as frames, feature maps and
attention matrices; upright bold lower-case symbols denote vector fields over the
image domain $\Omega$; italic symbols denote scalars, and upright subscripts are
descriptive labels.

\subsection{Physical Background and Rain-Flow Properties}
\label{sec:physics}  %

\textbf{Rain flow.}
A rain streak is the image-space trace of a falling droplet integrated over the
camera exposure. Let a droplet fall at terminal velocity $v_t$, be advected by
wind $\mathbf{w}$, and be observed by a camera with focal length $f$ and motion
$\mathbf{c}$. For a droplet at depth $Z(\mathbf{x})$, its projected image
velocity is
\begin{equation}
    \mathbf{u}(\mathbf{x})
    =
    \frac{f}{Z(\mathbf{x})}
    \left(
        v_t\hat{\mathbf{g}}
        +\mathbf{w}
        -\mathbf{c}
    \right)_{\perp},
\label{eq:rain_flow}
\end{equation}
where $\hat{\mathbf{g}}$ denotes the direction of gravity and
$(\cdot)_\perp$ extracts the image-plane component. Over an exposure of duration
$T$, the corresponding streak is oriented along $\mathbf{u}$ and has
approximate length $|\mathbf{u}|T$. We refer to $\mathbf{u}$ as the
\emph{rain flow}. It describes the rain structure within a frame and should not
be confused with optical flow, which describes scene correspondence between
frames.
The following three properties of rain flow form the physical basis of
FluidRain.

\textbf{Proposition 1 (Divergence-free rain flow).}
Under the incompressible-flow assumptions, the image-space rain flow satisfies
\begin{equation}
    \nabla\cdot\mathbf{u}=0.
\label{eq:div_free}
\end{equation}
For an arbitrary estimate $\hat{\mathbf{u}}$, its nearest admissible
divergence-free field in $L_2$ is
\begin{equation}
    \Pi(\hat{\mathbf{u}})
    =
    \hat{\mathbf{u}}-\nabla\phi,
    \qquad
    \nabla^2\phi=\nabla\cdot\hat{\mathbf{u}}.
\label{eq:projection}
\end{equation}

\textbf{Proposition 2 (Depth--scale equivariance).}
Let $D_s$ denote downsampling by a factor $s$. For fixed projected droplet
velocity, Eq.~\eqref{eq:rain_flow} gives
\begin{equation}
    \mathbf{u}_{sZ}
    =
    \frac{\mathbf{u}_{Z}}{s}
    =
    D_s[\mathbf{u}_{Z}]
    \qquad
    \text{in grid units}.
\label{eq:depth_scale}
\end{equation}
Hence, changing depth or image resolution rescales the apparent streak extent
while preserving its orientation.

\textbf{Proposition 3 (Short-term flow persistence and streak renewal).}
Over a short inter-frame interval, the underlying rain flow varies slowly,
whereas individual streak instances are not preserved across frames. Let
\begin{equation}
    \Delta\mathbf{u}_{\tau}
    =
    \mathbf{u}_{\tau+1}-\mathbf{u}_{\tau}.
\label{eq:flow_difference}
\end{equation}
Under the same assumptions, $\Delta\mathbf{u}_{\tau}$ remains bounded for
ordinary rain motion, while
rain-induced occlusions in neighbouring frames provide distinct observations of
the same scene.

\subsection{Rain-Flow Estimation and Attention Bias}
\label{sec:rain_flow}

\textbf{Rain-flow estimation.}
The rain flow in Eq.~\eqref{eq:rain_flow} depends on wind, scene depth, and
camera motion, which are unknown at test time. 
We therefore infer it directly from the observed rain streaks. Given the frame feature
$\mathbf{F}_\tau=\mathcal{S}(\mathbf{I}_\tau)$, a four-layer convolutional head
$\mathcal{H}$ predicts a raw rain-flow field.
\begin{equation}
    \hat{\mathbf{u}}_\tau=\mathcal{H}(\mathbf{F}_\tau).
\end{equation}
We predict $\hat{\mathbf{u}}_\tau$ at quarter resolution, which is sufficient
for capturing the slowly varying rain-flow structure while keeping the
estimator lightweight.
The same head $\mathcal{H}$ is reused for all input frames, so the flow estimator adds only 6,243 parameters.
 
However, $\hat{\mathbf{u}}_\tau$ is produced by an unconstrained neural
predictor and is not guaranteed to satisfy the divergence-free property in
Proposition~1. 
We therefore enforce this physical constraint by projecting the
raw estimate onto the divergence-free subspace.
\begin{equation}
    \mathbf{u}_\tau
    =
    \Pi(\hat{\mathbf{u}}_\tau).
\label{eq:estimated_flow}
\end{equation}
After projection, $\mathbf{u}_\tau$ serves as the rain-flow field for frame
$\mathbf{I}_\tau$. Its direction indicates the local rain-streak orientation, while its
magnitude sets the along-streak extent of the attention bias.
We next use these two quantities to construct the rain-flow attention bias.

\textbf{Rain-flow attention bias.}
Since rain flow characterizes rain-streak geometry rather than scene
correspondence, we do not use it for feature warping.
Instead, we convert its orientation and extent into an additive bias on the
attention logits, encouraging attention to follow the local rain structure.

Consider a $w\times w$ attention window at scale $s$.
Let $\mathbf{u}_w$ denote the mean rain flow over the window.
The local rain flow $\mathbf{u}_w$ determines both the orientation and extent
of the rain streak. At scale $s$, we define
\begin{equation}
    \hat{\mathbf{t}}
    =
    \frac{\mathbf{u}_w}{|\mathbf{u}_w|},
    \qquad
    \hat{\mathbf{n}}\perp\hat{\mathbf{t}},
    \qquad
    \ell_s
    =
    \max\left(\frac{|\mathbf{u}_w|}{s^{2}},1\right),
\label{eq:extent}
\end{equation}%
where $\hat{\mathbf{t}}$ and $\hat{\mathbf{n}}$ denote the directions along
and across the streak, respectively, and $\ell_s$ is its extent in grid units.
The grid-unit flow $\mathbf{u}_w/s$ is divided by $s$ once more, so
$\ell_s$ is a local extent of one to two cells for the trained models;
the direction is what is carried across scales.

These quantities directly parameterize the attention bias:
$\hat{\mathbf{t}}$ and $\hat{\mathbf{n}}$ specify the along-streak and
across-streak directions, while $\ell_s$ sets the effective range along the
streak.
For a query at $\mathbf{q}$ and a key at $\mathbf{k}$, let
$\mathbf{d}=\mathbf{k}-\mathbf{q}$. The additive rain-flow bias
$\mathbf{B}_{\mathrm{flow}}^{(h)}$ for attention head $h$ has entries
\begin{equation}
\big[\mathbf{B}_{\mathrm{flow}}^{(h)}\big]_{\mathbf{d}}
=
-\frac{4}{w^2}
\left[
a_h(\mathbf{d}\!\cdot\!\hat{\mathbf{n}})^2
+
b_h
\frac{(\mathbf{d}\!\cdot\!\hat{\mathbf{t}})^2}{\ell_s^2}
\right],
\qquad
a_h,b_h\geq0.
\label{eq:flow_bias}
\end{equation}
The two terms control the bias in the directions normal and tangent to the
rain streak, respectively. 
Displacements across the streak are penalized directly, whereas displacements
along the streak are normalized by $\ell_s^2$. Consequently, a larger
$\ell_s$ yields a broader attention range along the rain direction, producing
an anisotropic bias aligned with the local streak.
The bias adds only two learnable parameters, $a_h$ and $b_h$, for each attention head.

\subsection{Loop-in-Loop across Scale and Time}%
\label{sec:loop}
\label{sec:backbone}  %
\label{sec:diffbias}  %
A weight-tied operator is most useful when each reuse receives complementary
information rather than the same observation. 
According to Propositions~2 and~3, rain provides such variation across both scale and time: streak extent changes across resolutions, while neighboring frames contain renewed streak instances under a similar flow structure. This motivates our \emph{Loop-in-Loop} design, with an outer loop across scales and an inner loop across neighboring frames.

\textbf{Outer loop: sharing across scales.}
Proposition~2 shows that downsampling preserves the rain-flow direction and
rescales its magnitude by $1/s$.  Therefore, after expressing the
flow in the current grid as $\mathbf{u}/s$, the same rain-aware operator can be
applied at different resolutions. Sharing is motivated primarily by this
scale-consistent direction; the extent $\ell_s$ of Eq.~\eqref{eq:extent}
remains local at each scale. We thus reuse the same block stack
$\mathcal{B}$ at three scales $s\in\{1,2,4\}$.

For each scale $s$, the centre-frame feature is first resized as
\begin{equation}
    \mathbf{X}_s = D_s(\mathbf{X}),
\label{eq:scale_feature}
\end{equation}
where $D_s$ denotes resampling by a factor $s$. The corresponding rain flow is
represented as $\mathbf{u}_t/s$ and converted into the rain-flow bias using
Eq.~\eqref{eq:flow_bias}. The shared stack then produces
\begin{equation}
    \mathbf{Y}_s
    =
    \mathcal{B}\!\left(
        \mathbf{X}_s;
        \mathbf{B}_{\mathrm{flow}}(\mathbf{u}_t/s)
    \right),
\label{eq:outer_loop}
\end{equation}
where $\mathbf{Y}_s$ is the processed feature at scale $s$. After upsampling,
$\mathbf{Y}_s$ becomes the state $\mathbf{X}$ for the next scale; the three upsampled
outputs are fused with weights $\operatorname{Softmax}(\mathbf{W}_f\mathbf{M})$,
$\mathbf{M}$ being the clip descriptor defined below, and predict the residual added
to $\mathbf{I}_t$.

\textbf{Inner loop: sharing across frames.}
Proposition~3 shows that neighboring frames share similar rain-flow structure
while containing different rain-streak realizations, and thus provide
complementary observations for restoring the centre frame. We therefore use the
centre-frame state as the query and reuse the same attention operator to read
keys and values from the centre and neighboring frames without explicit
alignment.

At scale $s$, let $\mathbf{X}$ denote the current centre-frame state and
$\mathbf{N}_\delta$ the feature of frame $t+\delta$, where
$\delta\in\{0,-1,+1\}$. The centre state provides the query,
\begin{equation}
    \mathbf{Q} = \mathbf{X}\mathbf{W}_Q,
\end{equation}
while each temporal pass obtains its keys and values from the frame being read.
\begin{equation}
    \mathbf{K}_\delta = \mathbf{N}_\delta\mathbf{W}_K,
    \qquad
    \mathbf{V}_\delta = \mathbf{N}_\delta\mathbf{W}_V.
\end{equation}
The corresponding attention output is
\begin{equation}
\mathbf{A}_\delta
=
\operatorname{Softmax}
\left(
    \frac{\mathbf{Q}\mathbf{K}_\delta^\top}{\sqrt{d}}
    +
    \mathbf{B}_{\mathrm{rel}}
    +
    \mathbf{B}_{\mathrm{flow}}
    \left(
        \frac{\mathbf{u}_{t+\delta}}{s}
    \right)
\right)\mathbf{V}_\delta ,
\label{eq:temporal_attention}
\end{equation}
where $\mathbf{B}_{\mathrm{rel}}$ is the standard learned relative-position bias and
$\mathbf{B}_{\mathrm{flow}}$ is defined in Eq.~\eqref{eq:flow_bias}.
After reading the centre and two neighbouring frames with the shared
attention operator, we obtain three attention outputs
$\mathbf{A}_0$, $\mathbf{A}_{-1}$, and $\mathbf{A}_{+1}$.
Rather than combining them equally, we predict three fusion weights from
the centre-frame feature and a simple summary of the rain-flow change.
Specifically, with
$\Delta\mathbf{u}=\mathbf{u}_{t+1}-\mathbf{u}_t$, we define
\begin{equation}
    \psi(\Delta\mathbf{u})
    =
    \left[
        \overline{|\Delta\mathbf{u}|},
        \;
        \overline{\cos\angle(\mathbf{u}_{t+1},\mathbf{u}_t)}
    \right],
    \qquad
    \mathbf{M} = \mathbf{W}_m\overline{\mathbf{F}_t}+\mathbf{W}_d\,\psi(\Delta\mathbf{u}),
\label{eq:flow_descriptor}
\end{equation}
where $\mathbf{M}$ is a compact clip descriptor and $\psi$ summarizes the
frame-to-frame change in rain flow. The temporal fusion weights are then
obtained as
\begin{equation}
    g=\operatorname{Softmax}(\mathbf{W}_g\mathbf{M}),
    \qquad
    \mathbf{X}
    \leftarrow
    \mathbf{X}+
    \sum_{\delta\in\{0,-1,+1\}}
    g_\delta \mathbf{A}_\delta .
\label{eq:temporal_fusion}
\end{equation}
The fused feature is subsequently processed by the standard MLP sublayer,
completing the inner-loop update.
Together with the outer scale loop, this forms the Loop-in-Loop architecture:
the same block stack is reused across resolutions, while the same attention
operator is reused to read different frames within each block.

\subsection{Training}
\label{sec:loss}  %

The network is trained primarily with an $\ell_1$ reconstruction loss on the
restored centre frame:
\begin{equation}
    \mathcal{L}_{\mathrm{rec}}
    =
    \left\|
        \hat{\mathbf{I}}_t-\mathbf{I}_t^{\mathrm{gt}}
    \right\|_1 .
\end{equation}
On synthetic data, the rain layer
$\mathbf{R}_t=\mathbf{I}_t-\mathbf{I}_t^{\mathrm{gt}}$ also provides supervision for the rain-flow head.
For each quarter-resolution cell containing rain, we extract a target streak
orientation $\theta^\ast$ and length $\ell^\ast$. Let $\theta$ and $\ell$
denote the corresponding quantities from the projected flow $\mathbf{u}_t$.
The flow loss is
\begin{equation}
\mathcal{L}_{\mathrm{flow}}
=
\frac{1}{|\Omega_r|}
\sum_{\Omega_r}
\left[
    1-\cos 2(\theta-\theta^\ast)
    +
    \left(
        \log\ell-\log\ell^\ast
    \right)^2
\right],
\label{eq:flow_loss}
\end{equation}
where $\Omega_r$ denotes cells containing visible rain.
The total objective is
\begin{equation}
    \mathcal{L}
    =
    \mathcal{L}_{\mathrm{rec}}
    +
    \lambda_{\mathrm{flow}}
    \mathcal{L}_{\mathrm{flow}}.
\label{eq:total_loss}
\end{equation}
For NTURain we use $\lambda_{\mathrm{flow}}=0.01$, selected on the validation
set. For RainSynLight, RainSynComplex, and RainSynAll100,
$\lambda_{\mathrm{flow}}=0$, so the rain-flow head is learned solely through
the restoration objective.

\section{Experiments}
\label{sec:exp}

\begin{table}[t]
\caption{\textbf{Quantitative comparison on video deraining benchmarks.}
Best and second-best results are highlighted in bold and underline.}
\label{tab:main}
\centering
\scriptsize
\setlength{\tabcolsep}{1.25pt}

\resizebox{\textwidth}{!}{%
\begin{tabular}{l l r r ccc ccc ccc ccc}
\toprule
& & & &
\multicolumn{3}{c}{NTURain} &
\multicolumn{3}{c}{RainSynLight} &
\multicolumn{3}{c}{RainSynComplex} &
\multicolumn{3}{c}{RainSynAll100} \\
\cmidrule(lr){5-7}
\cmidrule(lr){8-10}
\cmidrule(lr){11-13}
\cmidrule(lr){14-16}
Method &
Venue &
\#P (M) &
Time (ms) $\downarrow$ &
PSNR $\uparrow$ &
SSIM $\uparrow$ &
LPIPS $\downarrow$ &
PSNR $\uparrow$ &
SSIM $\uparrow$ &
LPIPS $\downarrow$ &
PSNR $\uparrow$ &
SSIM $\uparrow$ &
LPIPS $\downarrow$ &
PSNR $\uparrow$ &
SSIM $\uparrow$ &
LPIPS $\downarrow$ \\
\midrule
SwinIR-light & ICCVW'21 & 0.91 & 803
& 32.79 & 0.952 & 0.0572
& 33.25 & 0.956 & 0.0667
& 26.61 & 0.848 & 0.2232
& 32.49 & 0.924 & 0.1150 \\

Restormer & CVPR'22 & 26.11 & 232
& 35.09 & \best{0.962} & \best{0.0355}
& 37.19 & \second{0.978} & \second{0.0241}
& 30.07 & 0.897 & \second{0.1302}
& \second{38.06} & \best{0.965} & \best{0.0417} \\

NAFNet & ECCV'22 & 29.16 & \best{20}
& 33.88 & 0.956 & \second{0.0432}
& 35.60 & 0.967 & 0.0443
& 28.04 & 0.854 & 0.1980
& 35.30 & 0.934 & 0.1007 \\

\midrule
IDT & TPAMI'23 & 16.42 & 249
& 33.23 & 0.952 & 0.0586
& 35.81 & 0.969 & 0.0404
& 27.97 & 0.860 & 0.1894
& 35.44 & 0.946 & 0.0733 \\

DRSformer & CVPR'23 & 33.66 & 468
& 34.07 & 0.958 & 0.0474
& 37.46 & 0.977 & 0.0244
& 29.68 & 0.892 & 0.1356
& 37.41 & 0.958 & 0.0549 \\

NeRD-Rain & CVPR'24 & 22.89 & 298
& 34.71 & 0.960 & 0.0444
& 37.47 & \second{0.978} & 0.0244
& 29.79 & 0.888 & 0.1404
& \best{38.16} & \second{0.959} & \second{0.0546} \\

\midrule
S2VD & CVPR'21 & 0.53 & \second{28}
& 34.17 & 0.956 & 0.0504
& 25.36 & 0.913 & 0.1389
& 28.23 & 0.836 & 0.2242
& 31.53 & 0.878 & 0.1637 \\

ESTINet & TPAMI'23 & 29.90 & 513
& \second{35.65} & \best{0.962} & 0.0450
& \best{40.74} & \best{0.983} & \best{0.0195}
& \best{31.48} & \best{0.912} & \best{0.1135}
& 35.85 & 0.945 & 0.0751 \\

RainMamba & ACM MM'24 & 30.86 & 75
& 32.65 & 0.942 & 0.0527
& 33.49 & 0.934 & 0.0816
& 27.75 & 0.809 & 0.2861
& 29.89 & 0.864 & 0.1993 \\

VDMamba & CVPR'25 & 12.70 & 168
& 34.00 & 0.953 & 0.0538
& 35.38 & 0.946 & 0.0946
& 27.85 & 0.806 & 0.2753
& 31.13 & 0.866 & 0.2047 \\

DeLiVR & ICLR'26 & 5.63 & 920
& 32.79 & 0.944 & 0.0714
& 33.41 & 0.941 & 0.0948
& 26.47 & 0.785 & 0.3199
& 32.26 & 0.872 & 0.1937 \\

\midrule
\textbf{\ours{}} & -- & 0.80 & 271
& \best{35.73} & \second{0.961} & 0.0446
& \second{37.52} & 0.973 & 0.0429
& \second{30.18} & \second{0.898} & 0.1782
& 35.64 & 0.948 & 0.0986 \\

\bottomrule
\end{tabular}
}
\end{table}

\subsection{Settings}

\textbf{Datasets.} We evaluate on NTURain \citep{chen2018robust}, RainSynLight and
RainSynComplex \citep{liu2018erase}, and RainSynAll100
\citep{yang2021recurrent}, covering synthetic and real rain, light and heavy
rain, non-linear background motion, and veiling effects. 
We further introduce \textbf{RainSyn-Gust} to evaluate robustness to changing
rain motion, which is absent from existing benchmarks. Since the rain layer is
exactly separable in RainSynLight and RainSynComplex, we rotate only the rain
layer and add it back to the clean frame, so that rain direction changes while
the background remains unchanged. We consider a $30^\circ$ step at the middle
frame (G-cam) and a $3^\circ$-per-frame ramp (G-wind).

\textbf{Protocol.} 
All methods are retrained under the same protocol on a single RTX 4090 GPU
and evaluated on the full test sets. We report PSNR and SSIM on synthetic
benchmarks, with LPIPS and FVD additionally reported on NTURain. For real
rainy videos without ground truth, we use our proposed physics-based
no-reference metric $F_{\mathrm{phys}}$. We compare with three groups of
baselines: general image restoration (SwinIR-light
\citep{liang2021swinir}, Restormer \citep{zamir2022restormer}, and NAFNet
\citep{chen2022nafnet}), single-image deraining (IDT
\citep{xiao2023idt}, DRSformer \citep{chen2023drsformer}, and NeRD-Rain
\citep{chen2024nerdrain}), and video deraining (S2VD
\citep{yue2021s2vd}, ESTINet \citep{zhang2022estinet}, RainMamba
\citep{wu2024rainmamba}, VDMamba \citep{sun2025vdmamba}, and DeLiVR
\citep{sun2026delivr}).

\subsection{Comparison on four benchmarks}%
\begin{figure}[t]
\centering
\includegraphics[width=\textwidth]{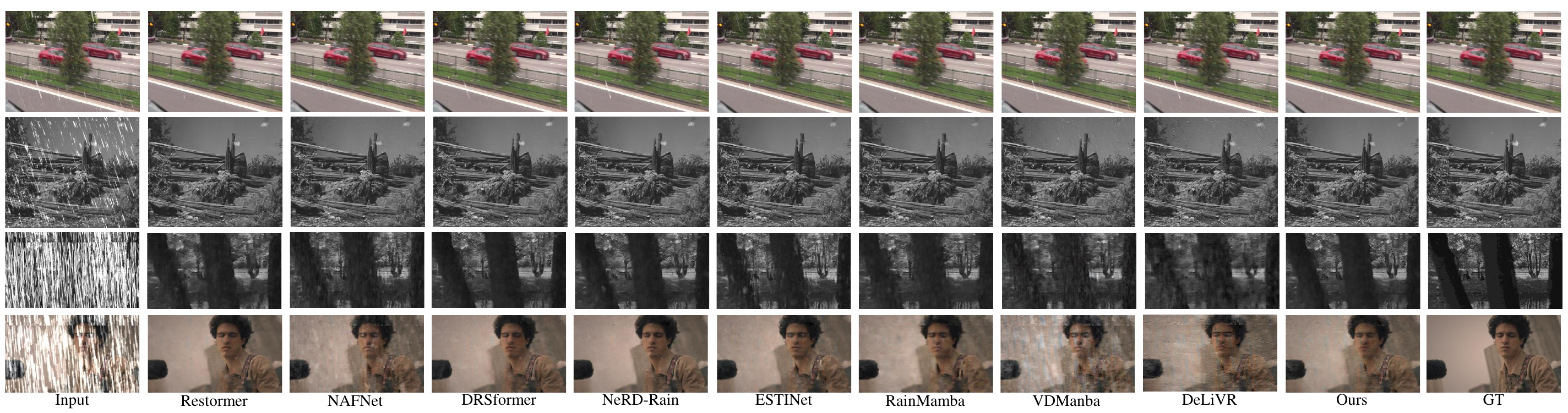}
\vspace{-5mm}
\caption{Qualitative comparison on video deraining benchmarks.}
\vspace{-3mm}
\label{fig:qual}
\end{figure}

Table~\ref{tab:main} compares FluidRain with representative image and video
deraining methods on four benchmarks. With only 0.80M parameters, FluidRain
remains competitive with substantially larger models, achieving the
second-highest PSNR on both NTURain and RainSynComplex while maintaining
strong performance on RainSynLight and RainSynAll100. These results indicate
that the proposed Loop-in-Loop design provides an effective trade-off between
restoration quality and model size. Qualitative comparisons in
Figure~\ref{fig:qual} further show that FluidRain effectively removes rain
streaks while preserving scene structures and fine details.

\subsection{Analysis of the Loop-in-Loop Design}
\label{sec:views}

Table~\ref{tab:views} examines where the benefit of Loop-in-Loop comes from.
For the view-based controls, we disable the rain-flow pathway and keep the
same backbone, changing only the observation available to each shared pass.
We additionally vary the temporal range and whether the scale-specific
operators share weights. The full FluidRain is included in the last column
for reference.

Simply repeating the centre-frame observation provides only marginal and
inconsistent gains, whereas changing the spatial view or reading a
neighbouring frame consistently improves restoration. The improvement becomes
substantially larger when both immediate neighbours are used, indicating that
the shared operator benefits primarily from complementary observations rather
than repeated computation on the same input. Extending the temporal range to
five frames does not provide a consistent advantage: it slightly improves
RainSynComplex but degrades NTURain and RainSynLight. Likewise, untying the
scale-specific operators increases the parameter count from 0.79M to 2.17M
without improving the shared three-frame configuration. These results support
using the weight-tied three-frame setting $\{t-1,t,t+1\}$ as the Loop-in-Loop
backbone. Adding the rain-flow pathway further improves this backbone on all
three benchmarks, with the largest gain on RainSynComplex.

\begin{table*}[t]
\centering
\caption{Analysis of the Loop-in-Loop design.
}
\label{tab:views}
\scriptsize
\setlength{\tabcolsep}{1.1pt}
\renewcommand{\arraystretch}{1.05}
\begin{tabular}{lccccccccc}
\toprule
&
\multicolumn{1}{c}{Baseline} &
\multicolumn{1}{c}{Repeat} &
\multicolumn{3}{c}{Complementary views} &
\multicolumn{2}{c}{Temporal range} &
\multicolumn{1}{c}{Weight Sharing} &
\multicolumn{1}{c}{Full model} \\
\cmidrule(lr){2-2}
\cmidrule(lr){3-3}
\cmidrule(lr){4-6}
\cmidrule(lr){7-8}
\cmidrule(lr){9-9}
\cmidrule(lr){10-10}

Setting
& Centre Only
& Repeat Centre
& Shifted Centre
& One neighbour
& Neighbour + shift
& Two neighbours
& Five frames
& Untied scales
& FluidRain \\

\midrule

Passes
& $t$
& $t,t$
& $t,t$
& $t,t{+}1$
& $t,t{+}1$
& $t,t{\pm}1$
& $t,t{\pm}1,t{\pm}2$
& $t,t{\pm}1$
& $t,t{\pm}1$ \\

\#P (M)
& 0.79
& 0.79
& 0.79
& 0.79
& 0.79
& 0.79
& 0.79
& 2.17
& 0.80 \\

\midrule

NTURain
& 32.57
& 32.73
& 33.06
& 33.54
& 33.78
& 34.55
& 34.25
& 33.81
& \textbf{35.73} \\

RainSynLight
& 33.29
& 33.17
& 33.79
& 35.65
& 35.66
& 36.85
& 36.71
& 36.11
& \textbf{37.52} \\

RainSynComplex
& 26.56
& 26.61
& 27.47
& 27.85
& 27.96
& 28.78
& 29.04
& 28.55
& \textbf{30.18} \\

\bottomrule
\end{tabular}
\vspace{-5mm}
\end{table*}

\subsection{Analysis of the Rain-Flow Pathway}
\label{sec:ablation}

\begin{table}[t]
\centering
\caption{Ablation of the rain-flow pathway.
}
\label{tab:ablation}
\small
\setlength{\tabcolsep}{3.8pt}

\begin{tabular}{lccccc}
\toprule
Variant
& \#P (M)
& NTURain
& RainSynLight
& RainSynComplex
& RainSynAll100 \\
\midrule

w/o rain-flow pathway
& 0.79
& 34.55
& 36.85
& 28.78
& 34.72 \\

\midrule

w/o projection
& 0.80
& 35.01
& 37.06
& 29.73
& 35.14 \\

w/o rain-flow bias
& 0.80
& 35.07
& 36.96
& 29.61
& 35.08 \\

w/o differential term
& 0.80
& 35.29
& 37.23
& 30.07
& 35.53 \\

\midrule

\ours{} (full)
& 0.80
& \textbf{35.73}
& \textbf{37.52}
& \textbf{30.18}
& \textbf{35.64} \\

\bottomrule
\end{tabular}
\vspace{-2mm}
\end{table}

Table~\ref{tab:ablation} evaluates the rain-flow pathway on top of the Loop-in-Loop backbone. 
With only 6.6K additional parameters, the full pathway
improves PSNR by 0.67--1.40 dB across the four benchmarks, with the largest gain on RainSynComplex. 
Removing either the divergence-free projection or the
rain-flow bias consistently degrades performance, whereas the differential term has only a minor effect. 
This suggests that the main benefit comes from
the projected flow and its use as directional attention guidance.
\begin{wrapfigure}{r}{0.48\linewidth}
    \centering
    \vspace{-6pt}
    \includegraphics[width=\linewidth]{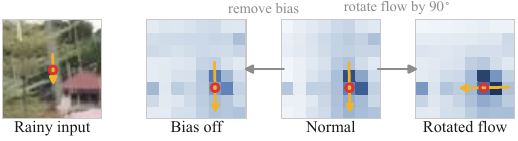}
    \vspace{-16pt}
    \caption{\textbf{Intervention on the rain-flow attention bias.}
    Removing the bias weakens the flow-aligned attention, while rotating
    the flow redirects it.}
    \label{fig:attn1}
    \vspace{-18pt}
\end{wrapfigure}
To understand how the rain-flow bias affects feature aggregation, we
intervene on the attention of a representative query in
Figure~\ref{fig:attn1}. With the learned bias, attention preferentially
extends along the projected rain-flow direction. Removing the bias weakens
this directional preference, while rotating the supplied flow by
$90^\circ$ redirects the attention accordingly. These interventions show
that the rain-flow bias explicitly steers the direction from which contextual
features are aggregated.

\subsection{Robustness to Changing Rain Motion}
\label{sec:gust}
\begin{figure}[t]
\centering
\includegraphics[width=\linewidth]{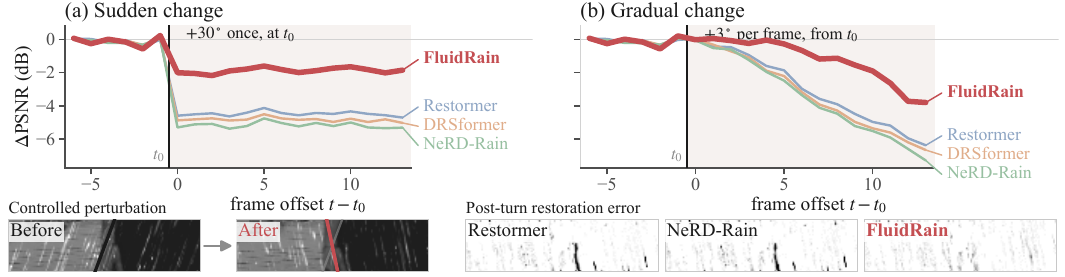}
\vspace{-5mm}
\caption{
\textbf{Robustness to changing rain motion on RainSyn-Gust.}
The bottom strip shows the controlled $30^\circ$ perturbation and the corresponding
post-turn restoration errors.
}
\vspace{-5mm}
\label{fig:gust}
\end{figure}

Existing benchmarks exhibit nearly constant rain directions within each sequence, leaving robustness to changing rain motion largely untested. 
We therefore construct \textbf{RainSyn-Gust} by rotating the separable rain layer while keeping the clean background fixed, with either a sudden $30^\circ$ change (\emph{G-cam}) or a gradual $3^\circ$-per-frame variation (\emph{G-wind}). 
As shown in Figure~\ref{fig:gust}, per-frame restoration
models suffer substantially larger relative PSNR degradation, whereas FluidRain remains stable under both settings. 
The rain-flow prior provides direction-aware guidance for aggregation, and Loop-in-Loop reuses the same operator across neighbouring frames and scales. 
The post-turn error maps further visualize the
improved robustness of FluidRain.

\subsection{$F_{\mathrm{phys}}$: No-Reference Evaluation on Real Rain}
\label{sec:phys}
Real rainy videos lack clean references, making PSNR and SSIM unavailable.
We therefore evaluate the removed residual
$\mathbf{r}_t=\mathbf{I}_t-\hat{\mathbf{I}}_t$ using $F_{\mathrm{phys}}$, which balances whether the removed content is physically consistent with rain (\emph{precision}) and how much rain-like content is removed (\emph{recall}).
\[
\mathrm{Prec}=(TCP)^{1/3},\qquad
\mathrm{Rec}=1-\frac{E_{\mathrm{rain}}(\hat{\mathbf{I}})}
{E_{\mathrm{rain}}(\mathbf{I})},\qquad
F_{\mathrm{phys}}
=\frac{2\,\mathrm{Prec}\,\mathrm{Rec}}
{\mathrm{Prec}+\mathrm{Rec}}.
\]
Here, $T$ measures temporal transience, $C$ streak coherence, and $P$ consistency with the divergence-free rain-flow constraint; $E_{\mathrm{rain}}(\cdot)$ measures rain-like residual energy.
\begin{figure}[h]
\centering
\includegraphics[width=\linewidth]{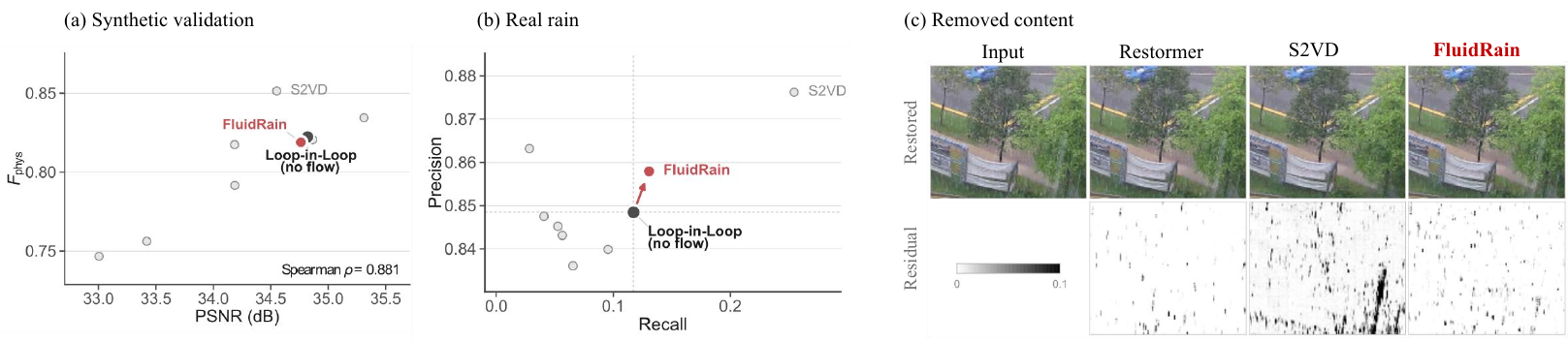}
\vspace{-5mm}
\caption{\textbf{Evaluation of $F_{\mathrm{phys}}$.}
(a) Correlation with PSNR on NTURain.
(b) Precision--recall on real rain.
(c) Restored crops and residuals.}
\vspace{-5mm}
\label{fig:real_residual}
\end{figure}

As shown in Figure~\ref{fig:real_residual}, $F_{\mathrm{phys}}$ agrees well with PSNR on synthetic NTURain (Spearman $\rho=0.881$), validating its use when references are unavailable. 
On real NTURain, FluidRain improves both precision and recall over its flow-less backbone, while the residual visualizations provide a qualitative view of the content removed by different methods.

\subsection{Downstream Perception under Rain}
\label{sec:downstream}
\begin{figure}[h]
\centering
\includegraphics[width=\linewidth]{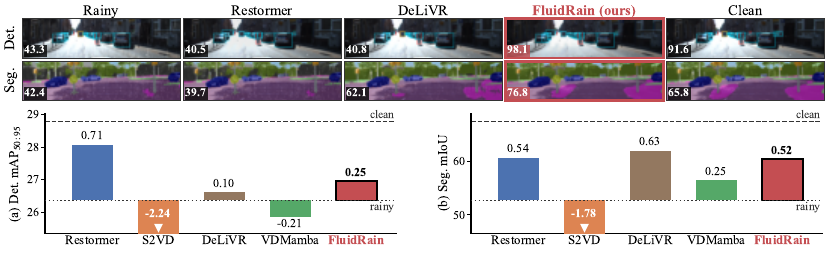}
\vspace{-5mm}
\caption{Downstream perception under rain.}
\vspace{-5mm}
\label{fig:downstream}
\end{figure}

We further evaluate the effect of deraining on downstream perception using
KITTI Tracking~\citep{geiger2012we} and KITTI 2015~\citep{menze2015object}.
As shown in Figure~\ref{fig:downstream}, \ours{} substantially improves both
detection and segmentation performance over the rainy inputs, recovering
$25\%$ of the detection degradation and $52\%$ of the segmentation
degradation. Despite having only $0.80$M parameters, \ours{} remains
competitive with substantially larger restoration models on both tasks.

\section{Conclusion}

We presented \ours{}, a lightweight video deraining framework built on
the physical structure of rain flow. FluidRain estimates a divergence-free
rain-flow field to guide attention and reuses the same operator across
resolutions and neighbouring frames through the Loop-in-Loop design.
This enables alignment-free restoration from three frames with only
0.80M parameters. Experiments on standard benchmarks, RainSyn-Gust, and
real rainy videos demonstrate the effectiveness and robustness of the
proposed design.

\subsubsection*{Reproducibility Statement}
Code, configuration files, and data-processing scripts will be released
with the paper. All reported tables and figures are
generated from stored experimental results using the released evaluation
scripts.

\subsubsection*{AI Use Statement}
Generative AI tools were used under the authors' direction to assist with
the development and refinement of hypotheses, experimental design,
software implementation, literature search, result interpretation, and
the drafting and editing of the manuscript. They were also used to assist
with mathematical derivations and proof writing. All AI-assisted code,
analyses, proofs, figures, and text were reviewed and verified by the
authors. The authors take full responsibility for the final content of
this work.

\bibliography{references_video}

@inproceedings{garg2004detection,
  title={Detection and removal of rain from videos},
  author={Garg, Kshitiz and Nayar, Shree K.},
  booktitle={CVPR},
  year={2004}
}

@article{garg2007vision,
  title={Vision and rain},
  author={Garg, Kshitiz and Nayar, Shree K.},
  journal={IJCV},
  volume={75}, number={1}, pages={3--27}, year={2007}
}

@inproceedings{yang2017jorder,
  title={Deep joint rain detection and removal from a single image},
  author={Yang, Wenhan and Tan, Robby T. and Feng, Jiashi and Liu, Jiaying and Guo, Zongming and Yan, Shuicheng},
  booktitle={CVPR},
  year={2017}
}

@inproceedings{guo2023sky,
  title={From sky to the ground: A large-scale benchmark and simple baseline towards real rain removal},
  author={Guo, Yun and Xiao, Xueyao and Chang, Yi and Deng, Shumin and Yan, Luxin},
  booktitle={ICCV},
  year={2023}
}

@article{kim2015video,
  title={Video deraining and desnowing using temporal correlation and low-rank matrix completion},
  author={Kim, Jin-Hwan and Sim, Jae-Young and Kim, Chang-Su},
  journal={IEEE TIP},
  volume={24}, number={9}, pages={2658--2670}, year={2015}
}

@inproceedings{chen2018robust,
  title={Robust video content alignment and compensation for rain removal in a {CNN} framework},
  author={Chen, Jie and Tan, Cheen-Hau and Hou, Junhui and Chau, Lap-Pui and Li, He},
  booktitle={CVPR},
  year={2018}
}

@inproceedings{liu2018erase,
  title={Erase or fill? {D}eep joint recurrent rain removal and reconstruction in videos},
  author={Liu, Jiaying and Yang, Wenhan and Yang, Shuai and Guo, Zongming},
  booktitle={CVPR},
  year={2018}
}

@article{yang2021recurrent,
  title={Recurrent multi-frame deraining: Combining physics guidance and adversarial learning},
  author={Yang, Wenhan and Tan, Robby T. and Feng, Jiashi and Wang, Shiqi and Cheng, Bin and Liu, Jiaying},
  journal={IEEE TPAMI},
  volume={44}, number={11}, pages={8569--8586}, year={2022}
}

@inproceedings{yue2021s2vd,
  title={Semi-supervised video deraining with dynamical rain generator},
  author={Yue, Zongsheng and Xie, Jianwen and Zhao, Qian and Meng, Deyu},
  booktitle={CVPR},
  year={2021}
}

@article{zhang2022estinet,
  title={Enhanced spatio-temporal interaction learning for video deraining: Faster and better},
  author={Zhang, Kaihao and Li, Dongxu and Luo, Wenhan and Ren, Wenqi and Liu, Wei},
  journal={IEEE TPAMI},
  volume={45}, number={1}, pages={1287--1293}, year={2023}
}

@inproceedings{yang2023viwsnet,
  title={Video adverse-weather-component suppression network via weather messenger and adversarial backpropagation},
  author={Yang, Yijun and Aviles-Rivero, Angelica I. and Fu, Huazhu and Liu, Ye and Wang, Weiming and Zhu, Lei},
  booktitle={ICCV},
  pages={13200--13210}, year={2023}
}

@inproceedings{wu2024rainmamba,
  title={{RainMamba}: Enhanced locality learning with state space models for video deraining},
  author={Wu, Hongtao and Yang, Yijun and Xu, Huihui and Wang, Weiming and Zhou, Jinni and Zhu, Lei},
  booktitle={ACM MM},
  pages={7881--7890}, year={2024}
}

@inproceedings{sun2025vdmamba,
  title={Semi-supervised state-space model with dynamic stacking filter for real-world video deraining},
  author={Sun, Shangquan and Ren, Wenqi and Zhou, Juxiang and Wang, Shu and Gan, Jianhou and Cao, Xiaochun},
  booktitle={CVPR},
  pages={26114--26124}, year={2025}
}

@inproceedings{sun2026delivr,
  title={{DeLiVR}: Differential spatiotemporal {L}ie bias for efficient video deraining},
  author={Sun, Shuning and Lu, Jialang and Chen, Xiang and Wang, Jichao and Lu, Dianjie and Zhang, Guijuan and Gao, Guangwei and Zheng, Zhuoran},
  booktitle={ICLR},
  year={2026}
}

@inproceedings{chan2022basicvsrpp,
  title={{BasicVSR++}: Improving video super-resolution with enhanced propagation and alignment},
  author={Chan, Kelvin C.K. and Zhou, Shangchen and Xu, Xiangyu and Loy, Chen Change},
  booktitle={CVPR},
  pages={5972--5981}, year={2022}
}

@article{xiao2023idt,
  title={Image de-raining transformer},
  author={Xiao, Jie and Fu, Xueyang and Liu, Aiping and Wu, Feng and Zha, Zheng-Jun},
  journal={IEEE TPAMI},
  volume={45}, number={11}, pages={12978--12995}, year={2023}
}

@inproceedings{chen2023drsformer,
  title={Learning a sparse transformer network for effective image deraining},
  author={Chen, Xiang and Li, Hao-Ran and Li, Mingqiang and Pan, Jinshan},
  booktitle={CVPR},
  year={2023}
}

@inproceedings{chen2024nerdrain,
  title={Bidirectional multi-scale implicit neural representations for image deraining},
  author={Chen, Xiang and Pan, Jinshan and Dong, Jiangxin},
  booktitle={CVPR},
  year={2024}
}

@inproceedings{liang2021swinir,
  title={{SwinIR}: Image restoration using {S}win transformer},
  author={Liang, Jingyun and Cao, Jiezhang and Sun, Guolei and Zhang, Kai and Van Gool, Luc and Timofte, Radu},
  booktitle={ICCVW},
  year={2021}
}

@inproceedings{liu2021swin,
  title={Swin transformer: Hierarchical vision transformer using shifted windows},
  author={Liu, Ze and Lin, Yutong and Cao, Yue and Hu, Han and Wei, Yixuan and Zhang, Zheng and Lin, Stephen and Guo, Baining},
  booktitle={ICCV},
  year={2021}
}

@inproceedings{zamir2022restormer,
  title={Restormer: Efficient transformer for high-resolution image restoration},
  author={Zamir, Syed Waqas and Arora, Aditya and Khan, Salman and Hayat, Munawar and Khan, Fahad Shahbaz and Yang, Ming-Hsuan},
  booktitle={CVPR},
  year={2022}
}

@inproceedings{chen2022nafnet,
  title={Simple baselines for image restoration},
  author={Chen, Liangyu and Chu, Xiaojie and Zhang, Xiangyu and Sun, Jian},
  booktitle={ECCV},
  year={2022}
}

@inproceedings{kim2016drcn,
  title={Deeply-recursive convolutional network for image super-resolution},
  author={Kim, Jiwon and Lee, Jung Kwon and Lee, Kyoung Mu},
  booktitle={CVPR},
  year={2016}
}

@inproceedings{dehghani2019universal,
  title={Universal transformers},
  author={Dehghani, Mostafa and Gouws, Stephan and Vinyals, Oriol and Uszkoreit, Jakob and Kaiser, {\L}ukasz},
  booktitle={ICLR},
  year={2019}
}

@article{garg2006photorealistic,
  title={Photorealistic rendering of rain streaks},
  author={Garg, Kshitiz and Nayar, Shree K.},
  journal={ACM TOG},
  volume={25}, number={3}, pages={996--1002}, year={2006}
}

@article{barnum2010analysis,
  title={Analysis of rain and snow in frequency space},
  author={Barnum, Peter C. and Narasimhan, Srinivasa and Kanade, Takeo},
  journal={IJCV},
  volume={86}, number={2--3}, pages={256--274}, year={2010}
}

@inproceedings{hu2019dafnet,
  title={Depth-attentional features for single-image rain removal},
  author={Hu, Xiaowei and Fu, Chi-Wing and Zhu, Lei and Heng, Pheng-Ann},
  booktitle={CVPR},
  year={2019}
}

@article{bossu2011rain,
  title={Rain or snow detection in image sequences through use of a histogram of orientation of streaks},
  author={Bossu, J{\'e}r{\'e}mie and Hauti{\`e}re, Nicolas and Tarel, Jean-Philippe},
  journal={IJCV},
  volume={93},
  number={3},
  pages={348--367},
  year={2011},
  publisher={Springer}
}

@inproceedings{bae2025relaxed,
  title={Relaxed recursive transformers: Effective parameter sharing with layer-wise {LoRA}},
  author={Bae, Sangmin and Fisch, Adam and Harutyunyan, Hrayr and Ji, Ziwei and Kim, Seungyeon and Schuster, Tal},
  booktitle={ICLR},
  year={2025}
}

@article{schwethelm2026much,
  title={How much is one recurrence worth? {I}so-depth scaling laws for looped language models},
  author={Schwethelm, Kristian and Rueckert, Daniel and Kaissis, Georgios},
  journal={arXiv preprint arXiv:2604.21106},
  year={2026}
}

@article{xue2025asf,
  title={{ASF-Net}: Robust video deraining via temporal alignment and online adaptive learning},
  author={Xue, Xinwei and He, Jia and Ma, Long and Meng, Xiangyu and Li, Wenlin and Liu, Risheng},
  journal={Pattern Recognition},
  volume={158},
  pages={110973},
  year={2025},
  publisher={Elsevier}
}

@inproceedings{lin2025controllable,
  title={Controllable weather synthesis and removal with video diffusion models},
  author={Lin, Chih-Hao and Wang, Zian and Liang, Ruofan and Zhang, Yuxuan and Fidler, Sanja and Wang, Shenlong and Gojcic, Zan},
  booktitle={ICCV},
  pages={13580--13591},
  year={2025}
}

@inproceedings{he2026vision,
  title={{Vision-MoR}: Scaling Vision Transformer via Patch-Level Mixture-of-Recursions},
  author={He, Yunhong and Yuan, Zhengqing and Sun, Weixiang and Li, Yiyang and Liu, Yixin and Ye, Yanfang and Sun, Lichao},
  booktitle={AAAI},
  volume={40},
  number={6},
  pages={4699--4707},
  year={2026}
}

@inproceedings{geiger2012we,
  title={Are we ready for autonomous driving? {T}he {KITTI} vision benchmark suite},
  author={Geiger, Andreas and Lenz, Philip and Urtasun, Raquel},
  booktitle={CVPR},
  pages={3354--3361},
  year={2012}
}

@inproceedings{menze2015object,
  title={Object scene flow for autonomous vehicles},
  author={Menze, Moritz and Geiger, Andreas},
  booktitle={CVPR},
  pages={3061--3070},
  year={2015}
}
\bibliographystyle{iclr2027_conference}

\end{document}